\documentclass{article}

\usepackage{microtype}
\usepackage{booktabs} % for professional tables

\usepackage{hyperref}

\usepackage[accepted]{icml2023}

\usepackage{amsmath}
\usepackage{amssymb}
\usepackage{mathtools}
\usepackage{amsthm}

\usepackage[capitalize,noabbrev]{cleveref}

\theoremstyle{plain}
\newtheorem{theorem}{Theorem}[section]

\theoremstyle{definition}
\newtheorem{definition}[theorem]{Definition}

\theoremstyle{remark}

\usepackage[textsize=tiny]{todonotes}

\usepackage[utf8]{inputenc} % allow utf-8 input
\usepackage[T1]{fontenc}    % use 8-bit T1 fonts
\usepackage{hyperref}       % hyperlinks
\usepackage{url}            % simple URL typesetting
\usepackage{booktabs}       % professional-quality tables
\usepackage{amsfonts}       % blackboard math symbols
\usepackage{nicefrac}       % compact symbols for 1/2, etc.
\usepackage{microtype}      % microtypography
\usepackage{xcolor}         % colors

\usepackage{enumitem}

\usepackage[normalem]{ulem}
\usepackage{xspace}

\usepackage{wrapfig}
\usepackage{multirow}
\usepackage{amsfonts, amsthm, amsmath, float}
\usepackage{bbm}
\usepackage{mathtools}
\usepackage{bm}
\usepackage{breqn}
\usepackage{tikz}
\usepackage{pgfplots}
\usepackage{scalefnt}
\usepackage{graphicx}
\usepackage{caption}
\usepackage{subcaption}
\usepackage[
  separate-uncertainty = true,
  multi-part-units = repeat
]{siunitx}

\usepackage{amssymb}
\usepackage{bbm}
\usepackage{mathtools}
\usepackage{bm}
\usepackage{algorithm} 
\usepackage{wrapfig}
\pgfplotsset{xticklabel={\tick},scaled x ticks=false}
\pgfplotsset{plot coordinates/math parser=false}
\DeclarePairedDelimiter\ceil{\lceil}{\rceil}
\DeclareMathOperator*{\argmax}{arg\,max}

\DeclareMathOperator{\ReLU}{ReLU}

\newcommand{\teal}[1]{\textcolor{teal}{#1}}
\newcommand{\golnoosh}[1]{\teal{\textsc{Golnoosh:} #1}}

\newcommand{\feta}{{{\sc{FETA}}}}

\icmltitlerunning{FETA: Fairness Enforced Verifying, Training, and Predicting Algorithms for Neural Networks}

\begin{document}

\twocolumn[
\icmltitle{FETA: Fairness Enforced Verifying, Training, and Predicting Algorithms for Neural Networks}




\begin{icmlauthorlist}
\icmlauthor{Kiarash Mohammadi}{mila,udem}
\icmlauthor{Aishwarya Sivaraman}{ucla}
\icmlauthor{Golnoosh Farnadi}{mila,udem,hec}
\end{icmlauthorlist}

\icmlaffiliation{mila}{Mila}
\icmlaffiliation{udem}{Université de Montréal}
\icmlaffiliation{ucla}{University of California, Los Angeles}
\icmlaffiliation{hec}{HEC Montréal}

\icmlcorrespondingauthor{Kiarash Mohammadi}{kiarash.mohammadi@mila.quebec}

\icmlkeywords{Machine Learning, ICML}

\vskip 0.3in
]



\printAffiliationsAndNotice{\icmlEqualContribution} % otherwise use the standard text.

\begin{abstract}
Algorithmic decision-making driven by neural networks has become very prominent in applications that directly affect people's quality of life. In this paper, we study the problem of verifying, training and guaranteeing individual fairness of neural network models. A popular approach for enforcing fairness is to translate a fairness notion into constraints over the parameters of the model. However, such a translation does not always guarantee fair predictions of the trained neural network model. To address this challenge, we develop a counterexample-guided post-processing technique to provably enforce fairness constraints at prediction time. Contrary to prior work that enforces fairness only on points around test or train data, we are able to enforce and guarantee fairness on all points in the input domain. Additionally, we propose an in-processing technique to use fairness as an inductive bias by iteratively incorporating fairness counterexamples in the learning process. We have implemented these techniques in a tool called {\feta}. Empirical evaluation on real-world datasets indicates that {\feta} is not only able to guarantee fairness on-the-fly at prediction time but also is able to train accurate models exhibiting a much higher degree of individual fairness.
\end{abstract}

\section{Introduction}
Deep neural networks are increasingly used to make sensitive decisions, including financial decisions
such as loan approval~\cite{hardt2016equality}, recidivism risk assessments~\cite{compas}, salary prediction~\cite{bbc}, etc. In these settings, for ethical, and legal reasons, it is of utmost importance that decisions are fair. For example, all else being equal, one would expect two individuals of a different gender to receive the same hiring decision. However, prior studies have shown that models trained on data are prone to bias on the basis of sensitive attributes such as race, gender, age, etc. \cite{propublica_compas, pmlr-v81-buolamwini18a} It has been shown that even if sensitive features such as race and gender
are withheld from the model, the model can still be unfair as it is often possible to reconstruct sensitive features that are encoded in data internally. Guaranteeing fairness not only helps organizations to address laws against discrimination but also helps users to better trust and understand the learned model~\cite{bastani2019probabilistic}.

Training neural network models such that fairness properties hold in their prediction is not always straightforward or possible. 
% A common approach to enforce fairness is though 
%Unfortunately, there is no easy way to enforce that a trained neural network should be fair in its sensitive features for all points in the input domain.
Existing approaches to the problem, either identify the absence of unfair predictions using verification~\cite{john2020verifying, urban2020perfectly} or guarantee fairness only for points during the training phase by constrained optimization techniques~\cite{choi2020learning} or add fairness regularizers to the loss function~\cite{kamishima2011fairness}. Other works which focus on test data points, provide fair models by % and test data points 
%\kia{todo: citation above?}
using robustness techniques~\cite{ruoss2020learning, yurochkin2019training}. While these techniques are successful in mitigating discrimination, they fail to provide global fairness guarantees for all points in the input domain at the prediction time. Over the past years, multiple definitions of fairness have been introduced. It is believed that none of these definitions dominates the others, and each of them is suitable for different settings. 
% that two population of individuals are treated equality on average. %Some of the well-known group fairness measures are statistical parity, equality of opportunity~\cite{hardt2016equality}.
%Finally, m
Recent works on fairness consider group-based notions of fairness~\cite{grarifairdecisiontrees, zafar2017fairness}, e.g., demographic parity~\cite{dwork2012fairness} or equalized odds~\cite{hardt2016equality}, that indicate that two populations of individuals should be treated equally on average. % Group fairness measures define specific groups in the population and require that particular statistics, computed based on model decisions, should be equal for all groups~\cite{hardt2016equality, dwork2012fairness}. 
Despite their prevalence, group fairness notions are generally hard to formally guarantee fairness for all input points~\cite{kearns2018preventing, ruoss2020learning}. Further, an algorithm that satisfies group fairness
could be blatantly unfair from the point of view of individual users~\cite{dwork2012fairness}. In this paper, we focus on individual fairness~\cite{dwork2012fairness} which states that the distance between the outcome for two individuals should be bounded according to the degree of their similarity.%, while in group fairness the focus is on ensuring that particular statistics computed based on model decisions, should be equal for all groups.

%\golnoosh{can we make this shorter?}
\paragraph{Our approach.} This paper develops techniques to detect, incorporate and guarantee {\em individual} fairness constraints for all points in the input space to a standard {ReLU} 
neural network without imposing further restrictions on the hypothesis space. These techniques leverage
recent work that employs automated theorem provers to formally verify the properties of neural networks. We focus on the individual fairness notion introduced by~\cite{galhotra2017fairness}. This notion says that a model is fair if, the decision of the model is the same for any two individuals with various combinations of sensitive attributes when nonsensitive attributes are fixed. To guarantee fairness, we present a counterexample-guided algorithm that detects and provably guarantees fairness at prediction time, given an arbitrary ReLU 
neural network. For any given model, our post-processing approach works by computing a majority decision for a group of individuals who share nonsensitive attributes on-the-fly via verification counterexamples. %Empirically we show that we guarantee fairness with little to no loss in model quality at a computational cost on the order of a few ms on standard datasets.
Furthermore, we propose a novel counterexample-guided algorithm to incorporate fairness during training. We identify individual fairness counterexamples on the training data, inducing additional supervision for training the network, and perform this process iteratively. We
have implemented our algorithms in a tool called “\textbf{F}airness \textbf{E}nforced Verifying, \textbf{T}raining, and Predicting \textbf{A}lgorithm” (\textbf{\feta}). Empirical evaluations on real-world benchmark datasets demonstrate the effectiveness of our solutions to train fair and accurate models, while provably guaranteeing fairness at the prediction time. Empirically, the two algorithms, when used in conjunction, enable better generalization while guaranteeing fairness.

\textbf{Main contributions.} Our key contributions are: 1) A practical individual fairness verification approach that detects discrimination through counterexamples given an arbitrary {ReLU} neural network (see Section~\ref{sec:verification}: CE-Fair Verification). 2) A counterexample-guided online algorithm that provably guarantees individual fairness at prediction time (see Section~\ref{sec:majority}: CE-Fair Prediction). 3) A counterexample-guided re-training algorithm that incorporates individual fairness during training (see Section~\ref{sec:training}: CE-Fair Training). 4) An end-to-end available implementation of our methods in an open-source tool called \feta, together with an extensive evaluation of real-world datasets (see Section~\ref{sec:predictionresults} and Section~\ref{sec:trainingresults}).
%\golnoosh{e first need to briefly explain our approaches in a high-level languages and then clearly list our contributions}
%We show that our counterexample-guided learning algorithm improves fairness of the learnt model after retraining. 

%Finally, we demonstrate that {\feta} outperforms baselines X and Y on X real-world benchmarks.

%\golnoosh{we can add the organizing paragraph if we have  space}
%\paragraph{Organization.} Section~\ref{sec:background} introduces our problem statement and notation.  Sections~\ref{sec:majority} and~\ref{sec:training} respectively describe our proposed algorithms: counterexample-guided fair prediction and counterexample-guided fairness enforced training. Experimental results in Section~\ref{sec:eval} demonstrate the potential of {\feta} on real-world benchmark datasets. Section~\ref{sec:relatedwork} reviews related work in learning fair  functions. We conclude and provide future directions in Section~\ref{sec:conclusion}.
\vspace{-15pt}
\section{Related Work}
%\paragraph{Fairness in Machine Learning} 
%There has been a large literature that device fairness specifications, including group fairness (e.g. demographic parity~\cite{dwork2012fairness}, equal opportunity~\cite{hardt2016equality}) and individual fairness~\cite{dwork2012fairness}.
%\textit{Group fairness} audits are done by comparing the classifier predictions for different demographic subgroups in the audit data set. For instance, a classifier satisfies the definition of \textit{statistical parity} \cite{dwork2012fairness} if the subjects in the protected group and the unprotected group have an equal probability of being assigned to the positive predicted class, e.g.~if credit card approval is equally probable for both females and males. Group fairness notions, specifically demographic parity is easy to work with, however, the authors of [3] highlighted its insufficiency as a fairness constraint. Further,~\cite{kleinberg2016inherent, chouldechova2017fair} show that imperfect predictors cannot simultaneously satisfy equal odds and calibration unless the groups have identical base rates, i.e. rates of positive outcomes. \textit{Individual fairness} states that the distance between the classification outcome for two individuals should be bounded according to the degree of their similarity. In this paper, we focus on individual fairness introduced by~\cite{galhotra2017fairness}, which says that a model is fair if, the decision of the model is the same for any two individuals with various combinations of sensitive attributes when nonsensitive attributes are fixed.

Our paper is a contribution to the extensive literature on fairness in machine learning. In this section, we will contextualize our work and compare it to existing techniques for mitigating and verifying fairness in machine learning.

\textbf{Fairness Mitigation Algorithms.} Methods that seek to introduce fairness into machine learning
systems broadly fall into one of three categories: \emph{pre-processing}, \emph{in-processing}, and \emph{post-processing}. %those that use data pre-processing before training, in-processing during training, and post-processing after training. 
In this paper, we propose both in-processing and post-processing techniques for individual fairness. Unlike prior work, our proposed in-processing technique does not depend on oracles for feedback on violations~\cite{gillen2018online, jung2019eliciting} and the fair training algorithm uses counterexamples rather than adversarial training~\cite{yurochkin2019training, yurochkin2020sensei} to enforce fairness. Further, adversarial training is approximate adversary examples, whereas we find exact counterexamples. While other works on post-processing focus on group fairness or other definitions of individual fairness~\cite{hardt2016equality, pleiss2017fairness, lohia2019bias, petersen2021post, agarwal2018automated}, we propose guaranteed predictions via post-processing using the individual fairness definition from~\cite{galhotra2017fairness}. Post-processing by~\cite{lohia2019bias} uses the same definition of individual fairness, but they support only a single binary sensitive attribute, while we can handle multiple binary/continuous sensitive attributes. Further, their approach learns a new ML model to be used in post-processing and does not provide formal guarantees of fair predictions for all points in the domain.

\textbf{Verifying machine learning systems and Adversarial Learning.} Prior work on machine learning verification can be classified into (i) verification using satisfiability modulo theory (SMT) or mixed-integer linear programming (MILP)~\cite{katz2017reluplex, tjeng2017evaluating}, and (ii) verification using convex relaxations~\cite{singh2019abstract, dvijotham2018dual}. Our proposed approach uses the MILP encoding from prior work~\cite{kiarash2021CFE} to build a practical individual fairness verification approach. Further, prior works on fairness verification have been proposed in the context of probabilistic programs~\cite{albarghouthi2017fairsquare, bastani2019probabilistic} or linear kernels~\cite{john2020verifying}. Recent works propose adversarially robust algorithms which can be divided into empirical~\cite{kurakin2016adversarial,madry2017towards} and certified defenses~\cite{wong2017provable, sinha2017certifiable, raghunathan2018certified, hein2017formal}. Specifically~\cite{xu2021robust, nanda2021fairness, yurochkin2020sensei} propose adversarial training-based algorithms for fairness.
We are closely related to these works, in that we carry out adversarial training using counterexamples. However, we differ in two ways. First, to the best of our knowledge, there is no related work in the adversarial robustness literature for ensuring individual fairness using the definition from~\cite{galhotra2017fairness}. Second, related work in adversarial training only ensures correctness in the neighbourhood of a training point, while we globally search for a counterexample and are able to discover long-range fairness violations.
\label{sec:relatedwork}
\vspace{-15pt}
\section{Preliminaries}
\label{sec:background}
%\golnoosh{Can we change the name of this section to Counterexample-Guided Fair Verification? In this way, we can introduce the case of neural network verification that can be used to audit neural networks by external auditors. We don't need to add any new experiments for this, because we show that we are able to find counterexamples for our data points. But we can first introduce the general model that returns pairs as counterexamples and then extend it to the approach that we use that given a data-point can find counterexamples. }

We begin by introducing some common notations. 
Let $\mathcal{X}$ be the input space consisting of $d$ features where $\mathcal{X} \equiv N \times S$ and $S$ denotes protected or sensitive features, and $N$ denotes remaining input features, and suppose that it is a compact finite subset $\mathcal{X}=[L, U]^d$ of $N \in \mathbb{R}^k$ and $S \in \mathbb{N}^{d-k}$. Let $\mathcal{Y} \in \{0,1\}$ be the output space. We consider a supervised binary classification task, where $f_{\theta}: \mathcal{X} \rightarrow [0, 1]$ outputs a probability distribution over classes, $\theta$ denotes the parameters and the classifier $g_{\theta}: f_{\theta}(\mathcal{X} ) \rightarrow \mathcal{Y}$ 
is defined as $g_{\theta}:= \mathbbm{1}(f_{\theta}(\bm{x}) \geq \Delta)$ where $\Delta$ is some classification threshold and it assigns an input to a category identified by numeric code $\mathcal{Y}$. Let $D$ = $\{(\bm{x}_1,y_1), \dots, (\bm{x}_M,y_M)\}$ be the training dataset containing $M$ samples with $\bm{x}_m$ and $y_m$ respectively denoting the $m$th individual and the corresponding output. The most commonly used \emph{Binary Cross Entropy~(BCE)} loss for this task is:
{\footnotesize\begin{equation}
    \label{eq:loss}
    L_{BCE} (f_\theta) = -\frac{1}{M} \sum_{i=1}^{M} y_i \log(f_\theta(\bm{x}_i)) + (1-y_i) \log(1 - f_\theta(\bm{x}_i))
\end{equation}}
where the goal is to find the best $\theta$ that minimizes $L_{BCE}$ across the data-generation distribution rather than just over the finite $D$.

%\golnoosh{Can you fix this: not all features are real values $\mathbb{R}^d$}
Our goal will be to verify, guarantee and train an \emph{individually fair} model in some sensitive input features. We refer to the \emph{Causal Discrimination} definition, a notion of individual fairness proposed by~\cite{galhotra2017fairness}. 
%\golnoosh{sensitive and non-sensitive features should be defined above}
\begin{definition}\label{def:fairness} (Causal Discrimination) Assume a function $g_{\theta}: f_{\theta}(\mathcal{X})\rightarrow \mathcal{Y}$, %where $\mathcal{X} \equiv N \times S$ and $S$ denotes protected or sensitive features, and $N$ denotes remaining input features, 
such that $\mathcal{X}[1 \dots k] \in N$ and $\mathcal{X}[k+1 \dots d] \in S$. We define $g_{\theta}$ to be \emph{individually fair} in sensitive features $S$ iff for any two points $\bm{x}$, $\bm{x}' \in \mathcal{X}$ where $\bm{x}[i] = \bm{x}'[i], ~\forall i \in \{1 \dots k\}$, we have that $g_{\theta}(\bm{x})=g_{\theta}(\bm{x}')$. %If this holds, consequently, $f_{\theta}$ is also fair.
\end{definition}

%\kia{changed the definition}

%\golnoosh{here k is the size of the non-sensitive features and d is the size of the input space}
%\golnoosh{Let's mention that we focus on ReLU in this paper and the we can say in the rest of this paper we refer to them as neural networks in short. }

In Neural Networks~(NN), various nonlinear activation functions have been introduced. Among those, $\ReLU$ has been used widely and generalized well~\cite{glorot2011deep,xu2016deep,dos2019deep}, particularly in the context of verification~\cite{katz2017reluplex, huang2017safety} and robustness. %$\ReLU$ Neural Networks~(NN) generalize well and are widely used~\cite{glorot2011deep,xu2016deep,dos2019deep}, particularly in the context of verification~\cite{katz2017reluplex, huang2017safety} and robustness. 
Hence, we will assume $f_{\theta}$ is a $\ReLU$ neural network. Formal properties of neural networks are often verified by encoding the semantics of neural networks ($f_{\theta}$) as
logical constraints. While several approaches to encoding neural networks for verification have been studied \cite{nnVerifSurvey, unified}; we use the encoding and optimization approach similar to~\cite{kiarash2021CFE} that propose significantly faster techniques for our use-case than the ones relying on Satisfiability Modulo Theories. This is crucial to our goal of guaranteeing fairness at prediction time. %\kiarash{Addressed golnoosh's comment}

%\golnoosh{the relation between $\theta$ and $w$ and $b$ should be defined. maybe there is a better way to define MILP encoding for relu networks based on $\theta$, e.g., can we simply say $z_i = \theta_i.\hat{z_{i-1}}$}

%\kiarash{isn't it just $\theta = (\bm{W}, \bm{b}$)? We have $(\bm{W}, \bm{b}$) in the MIP encoding as well. Also, $z_i$s are not parameters but variables}

%\golnoosh{Yes, so can we either say: $\theta = (W,B)$ and for $W_i$ and $b_i$ respectively denoting weights and bias of $i$th layer or simply ignore $W$ and $b$ and say $z_i = \theta_i.\hat{z_{i-1}}$?}

%\golnoosh{n is the number of layers here}
\begin{definition}\label{def:nnencoding} (MILP Encoding of Neural Networks) Let $f_{\theta}$ be an $n$-layer fully-connected ReLU neural network with a single output where $\theta = (\bm{W},\bm{B})$ and $\bm{W}_i$ and $\bm{b}_i$ respectively denoting weights and bias of $i$th layer. The width of each layer is represented by $t_i$, the values of neurons before applying ReLU are represented by vector $\boldsymbol{z}_i, ~\forall i \in \{0 \dots n\}$ ($\bm{z}_0$ being the input), and their values after ReLU by $\hat{\boldsymbol{z}}_i, ~\forall i \in \{1 \dots n\}$,~\cite{mip_nn} proposes the following MILP encoding, $\forall i \in \{1 \dots n\}$:
\end{definition}
%Given an $n$-layer fully-connected ReLU neural network $f_{\theta}$ with a single output where the width of each layer is represented by $t_i$, the values of neurons before applying ReLU is represented by vector $\boldsymbol{z}_i, ~\forall i \in \{0 \dots n\}$ ($\bm{z}_0$ being the input), and their values after ReLU by $\hat{\boldsymbol{z}}_i, ~\forall i \in \{1 \dots n\}$,~\cite{mip_nn} propose the following MIP encoding, $\forall i \in \{1 \dots n\}$:
%\vspace{-10pt}
{\footnotesize\begin{subequations} \label{eq:mip}
    \begin{equation} \label{eq:mip_a}
        \boldsymbol{z}_{i} = \boldsymbol{W}_{i}\hat{\boldsymbol{z}}_{i-1} + \boldsymbol{b}_{i}
    \end{equation}
    %\vspace{-0.5cm}
    \begin{equation} \label{eq:mip_b}
    \begin{split}
        \boldsymbol{\delta}_i \in \{0, 1\}^{t_i}, \quad \hat{\boldsymbol{z}}_i \geqslant 0, \quad \hat{\boldsymbol{z}}_i \leqslant \boldsymbol{u}_i \cdot \boldsymbol{\delta}_i,\\
        \hat{\boldsymbol{z}}_i \geqslant \boldsymbol{z}_i, \quad \hat{\boldsymbol{z}}_i \leqslant \boldsymbol{z}_i - \boldsymbol{l}_i \cdot (1 - \boldsymbol{\delta}_i)
    \end{split}
    \end{equation}
\end{subequations}}
Equation~\eqref{eq:mip_a} encodes the linear relationship, while Equation~\eqref{eq:mip_b} encodes the ReLU activation function, i.e., $\hat{\boldsymbol{z}} = ReLU(z) = \max(0,z)$. $\boldsymbol{\delta}_i$ is a vector of binary variables representing the state of each ReLU as \emph{non-active} or \emph{active}. This encoding relies on bounds on the values of neurons, $\boldsymbol{l}_i, \boldsymbol{u}_i$. These bounds are computed using a linear approximation of the network proposed by \cite{planet}, given the bounds on input $\bm{l}_0 = L, \bm{u}_0 = U$. Moreover, in this work, we assume both continuous and discrete domains over the input variables, hence the \emph{mixed-integer linear program}~(MILP).

Formal properties of functions are often characterized in terms of their counterexamples, e.g.,~\cite{solar2006combinatorial,clarke2000counterexample,sivaraman2020counterexample}. %Counterexample-guided algorithms are prevalent in the field of formal methods, for example to verify~\cite{clarke2000counterexample} and synthesize programs~\cite{solar2006combinatorial}. Recently, these counterexample-guided techniques have been used to guarantee monotonic predictions~\cite{sivaraman2020counterexample}.
The techniques proposed in this paper will be centred around using counterexamples to the fairness specification. Counterexample-guided algorithms rely on the ability to \emph{find} counterexamples, which require that both the counterexample specification and the object of interest ($f_{\theta}$) to be encoded in a language amenable to automated reasoning. Definition~\ref{def:nnencoding} provides such an encoding for $f_{\theta}$. In the next section, we show how to add fairness constraints per Definition~\ref{def:fairness}, to identify counterexamples. 
\vspace{-4pt}
%\section{Counterexample-Guided Fair Verification~(CE-Fair Verification)}
\section{CE-Fair Verification}
\label{sec:verification}

In 2019, Apple launched a credit card application that was accused of gender bias~\cite{bbc-apple}. The scandal went viral when a couple who shared all of their bank accounts, assets, and credit cards, received different credit limits while only their gender was different in their applications. Inspired by this real-world example, in this section, we propose an approach that focuses on auditing and verifying a trained neural network model to detect such discriminatory outcomes. %, i.e.~evaluating whether a NN classifier satisfies the individual fairness in Definition~\ref{def:fairness}. 
We envision scenarios where the classifier is a proprietary model, belonging e.g.~to a company or a bank, and an external party wants to inspect the model to ensure that it is operating fairly. We introduce \emph{Counterexample-guided fair (\textsc{CE-Fair}) verification} that can identify these fairness violations.

%Following the fairness notion in Definition~\ref{def:fairness} and neural network encoding introduced in Definition~\ref{def:nnencoding}, we introduce \emph{Counterexample-guided fair verification} that can identify counterexamples to this specification by solving an optimization problem. 
%\aish{we need to show what optimization problem we are solving?}
% The process of CE-Fair \emph{verification} refers to the process of checking that the output of $f_{\theta}$ satisfies individual fairness using Definition~\ref{def:fairness_cg} for all choices of the input. 

%\golnoosh{One of the major comments of the previous submission was about the confusion around the objective or the exact optimization program for verification and how we find CE, we need to make it clear in this submission}
%\kia{We have actually explained here what MIP constraints are added to implement the CE-Fair Verification function (equation 3a and 3b)}
\begin{definition}(\textsc{CE-Fair} Verification)
\label{def:fairness_verification}
Consider example $x \in \mathcal{X}$, function $f_{\theta}: \mathcal{X} \rightarrow [0, 1]$ and sensitive features $S$ and non-sensitive features $N$. Then a \emph{fairness counterexample} for example $x$, function $f_{\theta}$, and $S$ is $x'$ such that (i) $\bm{x}[i] = \bm{x}'[i], ~\forall i \in \{1 \dots k\} \in N$, and (ii) $g_{\theta}(\bm{x}) \neq g_{\theta}(\bm{x}')$.
We then define the \textsc{CE-Fair} verification function $v$ for sensitive feature set $S$ that takes as input a function $f_{\theta}: \mathcal{X} \rightarrow [0, 1]$ and $x \in \mathcal{X}$ as follows:
\vspace{-15pt}
\end{definition}
{\footnotesize\begin{equation*}
    v(f_{\theta}, x) = 
    \begin{cases}
    x' \;\; \text{where $x'$ is the fairness counterexample}\\ 
    \phi \;\; \text{if no fairness counterexample exists}\\ 
    \end{cases}
\end{equation*}}
\vspace{-10pt}
\textsc{CE-Fair} verification function $v(f_{\theta}, x)$ can find counterexample $x'$ by solving the optimization problem in which the two kinds of constraints defined in Equation~\ref{def:fairness_verification} are added to the MILP encoding from Equation~\ref{eq:mip}. The feasible set of this optimization problem is explored using an optimizer backend and if there exists a solution satisfying these constraints, we will have a fairness counterexample. Formally, the following constraints will be added to the MILP formulated in Equations~\ref{eq:mip_a},~\ref{eq:mip_b} to encode fairness counterexamples:
{\footnotesize\begin{subequations} \label{eq:ce}
    \begin{equation} \label{eq:ce_a}
        \bm{z}_{0}[i] = \bm{x}[i], ~\forall i \in \{1 \dots k\}
    \end{equation}
    \vspace{-0.4cm}
    \begin{equation} \label{eq:ce_b}
        \mathbbm{1}(z_n \geq \Delta) = 1 - g_{\theta}(\bm{x})
    \end{equation}
\end{subequations}}
%\golnoosh{why z has two index here? while in the previous definition it only has one?}
%\aish{fixed the indexing issue, so z is a vector which is index by layer, so it is supposed to be z\_0[i] indicating layer 0 neuron i}
%\golnoosh{k is the number of layers here?}
%\aish{here k is the index of non-sensitive attributes, i agree this is not clear, same issue in defintion 3, see comment above}
where $\bm{z}_{0}[i]$ is the variable associated with the $i$-th neuron in layer 0 (input layer). Concretely, this fairness verification approach searches for a counterexample with the same nonsensitive features as $\bm{x}$ and any assignments to sensitive features ($\bm{z}_{0, i}$ where $i \in \{k+1 \dots d\}$), constraining the output of the model to be opposite to $g_{\theta}(\bm{x})$. Here we only search for a counterexample violating fairness as a constraint by adding the following as an objective to the MILP:
{\footnotesize{\begin{equation} \label{eq:vio_obj}
    |z_n - f_{\theta}(\bm{x})|
\end{equation}}}

However, we can also search for the counterexample with maximum violation (see Definition~\ref{def:maxverification}). 
%\golnoosh{here we also need to specify the search for violation vs maximum violation which comes later and refer to definition 6.1}
While \textsc{CE-Fair} verification is sufficient to audit and verify a trained model to identify counterexamples as per Definition~\ref{def:fairness}, in the case where there are counterexamples --which is often the case-- it is not clear how to \emph{guarantee} fairness, or how to enforce it during training. The next two sections present the counterexample-guided algorithms that address these challenges.

%\golnoosh{can we claim that extending our approach to multi-class classification (one vs others) and regression tasks, e.g., predicting credit limit, is straightforward? can we add a few sentences about it here and then if needed for the training section?}
%\kiarash{Yes, we can. I will add a sentence or two about it.}

%\paragraph{\textbf{Extension.}} Note that extending our approach to multi-class classification
%is straightforward. We only need to (adopt one-vs-rest approach) for each class is straightforward To extend our approach to regression, e.g., to predict credit limit, we need to modify Equation~\ref{eq:ce_b} to encode fairness counterexamples as $|f_{\theta}(x) - f_{\theta}(x^\prime)|>\epsilon$ where $\epsilon$ is a hyperparamter that needs to be defined based on the context.
\vspace{-4pt}
%\section{Counterexample-Guided Fair Predictions~(CE-Fair Prediction)}
\section{CE-Fair Prediction}
\label{sec:majority}

For ethical or legal reasons, a company may want to ensure that their model outcome is the same for all individuals irrespective of their sensitive attributes. Further, they may want to ensure fair predictions using techniques that do not require modifying and re-training the model under study. Such requirements can be due to the utilization of outsourced models without having any access to the original training data or having limited resources to re-train the model. In this section, we leverage our counterexample-guided verification approach to provably guarantee individual fairness at prediction time without any requirements to re-train the model. We propose an \emph{online} technique that leverages counterexamples to Definition~\ref{def:fairness} to construct fair predictions on-the-fly at prediction time.

A naive approach to guarantee fair predictions would be to return the same output for all individuals, e.g., the most frequent label in the training set. While this satisfies individual fairness, it leads to poor model performance (see Table~\ref{tab:ce-fair-predictions} in Section~\ref{sec:predictionresults}). However, this gives us intuition to %a better approach, instead of returning the majority decision for all individuals in the domain, we could 
return the majority decision for a group of individuals who share nonsensitive attributes. We define \textsc{CE-Fair} prediction that produces individually fair output for a given input $x$ and $f_{\theta}$ as:
% More precisely, we apply function $h$ to the output of $f$ to obtain 

%\golnoosh{It is not clear how h is used for prediction in algorithm 1}
%\kiarash{Algorithm 1 is implementing $h$ itself. I edited line 2 of the algorithm. How's that?}

\begin{definition} (\textsc{CE-Fair} Prediction) For an example $x\in \mathcal{X}$, function $f_{\theta}: \mathcal{X} \rightarrow [0, 1]$, we define a post-processing \textsc{CE-Fair} Prediction function $h$ such that:
{\footnotesize \begin{equation}
\label{eq:fair_pred}
    h \left(f_{\theta}(\boldsymbol{x}) \right) = \mathbbm{1} \left( \left(\sum_{\bm{x}' \in A(\bm{x})} g_{\theta}(\bm{x}') - \mathbbm{1}{\left(g_{\theta}(\bm{x}') = 0\right)} \right) \geqslant 0 \right)
\end{equation}}
\noindent where:
{\footnotesize \begin{equation}
\label{eq:assingment}
\begin{split}
    A(\bm{x}) := \{X \mid X[1]=\bm{x}[1], \dots, X[k]=\bm{x}[k],\\ X[k+1]=a_{k+1}, \dots, X[d]=a_d;\\ \forall a_{k+1}, \dots,a_d \in [L'_{k+1, \dots, d}, U'_{k+1, \dots, d}]^{d-k} \}
\end{split}
\end{equation}}

%\golnoosh{What is $A(x)$, $k$ and $d$ and should we change $X$ to $\mathcal{X}$?}
%\kia{Basically, $A(\bm{x})$ takes $\bm{x}$ as an input, and returns the set of datapoints which share the same nonsensitive features with $\bm{x}$ but vary in sensitive features, i.e., the set of all assignments to the sensitive features given fixed nonsensitives of $\bm{x}$. I agree that this is not the best way to express this. This was the best way I came up with for the workshop version but I'll think about it more.}
%\golnoosh{thank you Kiarash! we need to make sure that there is consistency in all the notions that we define and also the story that we are telling. I think you can connect this definition to the CE-Fair verification one. I suggest trying to change both at the same time and see if you can find a way to present both with similar notions}
\end{definition}
Basically, this is contrasting the $0$ and $1$ outputs of $g_{\theta}$ in the space of sensitive features; the counter increments when $g_{\theta}(\bm{x}^\prime)$ is $1$ and decrements otherwise. %In the credit assignment example, this would result in the same credit amount for male and female individuals with the same credit history.
\begin{theorem}
For any function $f_{\theta}$ and for any input $\boldsymbol{x} \in \mathcal{X}$ with $S$ as sensitive features, $h(f_{\theta}(\bm{x}))$ is individually~fair~in~$S$.
\end{theorem}
\vspace{-0.5cm}
\begin{proof}
The proof is trivial: $h$ outputs the same decision for all points within the group of all assignments to the sensitive attributes given fixed nonsensitive attributes of $\bm{x}$, thus, no fairness counterexample exists.
\end{proof}
%\golnoosh{I don't like the current format that we define two approaches where the first one is a naive approach. Instead, I suggest mentioning the naive approach and the complexity as a motivation to introduce our approach. We can formally define counterexample-guided counting using verification method.}
%\kiarash{I suppose you kindly edited this yourself, thanks!}
%\golnoosh{can we make this section much shorter and explain the time complexity with $O(d \times |S|)$ where $S$ is the size of sensitive attributes and $d$ is the domain size of each sensitive attribute? }
So far we have established a way to guarantee fair predictions for all input points based on the majority decision captured in function $h$. To identify the majority decision, %we propose the following two ways that counts the frequency of each label within the given group of assignments specified by fixing the nonsensitive features in $\bm{x}$. %\golnoosh{these sections are very wordy and we should make them much shorter to save space for re-training part}
%\textbf{(i) Counting by Enumeration.} % This is the straight-forward implementation of function $h$ \eqref{eq:fair_pred}. 
%The first approach to computing majority decision would be 
the simple approach is to enumerate all possible assignments of sensitive attributes. %for a fixed set of nonsensitive attributes. 
Concretely, given a test point $\boldsymbol{x}$, we could traverse all possible assignments to the sensitive features, counting the frequency of each label. %We stop when we have found $\ceil*{\frac{|A(\boldsymbol{x})|}{2}}$ assignments that result in a specific label, assuring that this is the majority decision. 
The computational complexity of this approach grows with the size of sensitive attributes and the domain size of each sensitive attribute. %, i.e., $O(d \times |S|)$ where $S$ is the size of sensitive attributes and $d$ is the domain size of each sensitive attribute. %Therefore, exhaustively enumerating even $\ceil*{\frac{|A(\boldsymbol{x})|}{2}}$ will increase prediction time significantly. 
This approach, however, is not practical and increases prediction time significantly (See Appendix). This motivates the next approach in which we compute $h$ and identify the majority decision by leveraging our MILP framework to find counterexamples. %The key idea is to count if there are $\ceil*{\frac{|A(\boldsymbol{x})|}{2}}$ number of counterexamples for the MIP encoding of $f_{\theta}$ constrained to $\boldsymbol{x}$.

%\golnoosh{this definition needs more work, this should be connected to Definition 4 in the previous section and a new MIP encoding should be added here that define the "findSolutions" function in the algorithm to show how these two are connected.}
%\golnoosh{Here define "FairCounterExampleCounting"}
%\kia{I'm not sure how to fix this definition. It was a concern of reviewers.}
\begin{definition} (\textsc{CE-Fair} Counting)
\label{def:lazcouting}
All counterexamples of a test sample $\boldsymbol{x}$ (i.e., $\mathcal{S}$) can be determined in an iterative way by adding the following constraints to the verification problem in Definition~\ref{def:fairness_verification} and solve in iteration $K+1$ as:
{\footnotesize\begin{subequations}
\label{eq:counting}
\begin{align}
&  x^\prime  = v(f_{\theta},x) & \\
& \sum_{i=1}^d(\sum_{s \in \mathcal{S}:x^{\prime^k}_s[i]=0} x^\prime[i] +  \sum_{s \in \mathcal{S}: x^{\prime^k}_s[i]=1} (1 - x^\prime[i] )) \geq 1  &	\\
& \quad		   k=1,\ldots,K   \label{eq:NoGood}
\end{align}
\label{eq:ceset}
\end{subequations}}
where $S$ is a set of all counterexamples of $x$ and $x{\prime^k_s}$ is a counterexample of $x$ which is included in the set $S$ from the previous iteration $k$. To satisfy Constraints~\ref{eq:NoGood}, the solution must differ in at least one entry for each $x^{\prime^k}$. Once  Problem~\ref{eq:counting} becomes infeasible, then all counterexamples have been determined. Since there is a finite number of feasible assignments (e.g., Equation~\ref{eq:assingment}), this iterative method will stop in a finite time.
%Given a test sample $\boldsymbol{x}$, using the optimization backend \cite{gurobi}, we explore the MIP search tree in pursuit of $\ceil*{\frac{|A(\boldsymbol{x})|}{2}}$ counterexamples (rather than only one) where their labels is opposite to that of $f_{\theta}(\boldsymbol{x})$. 
\end{definition}

The lazy constraints generation approach defined in Definition~\ref{def:lazcouting} can be implemented more efficiently by using the optimization backend \cite{gurobi}. Hence, instead of iteratively finding counterexamples, we explore the MILP search tree in pursuit of $\ceil*{\frac{|A(\boldsymbol{x})|}{2}}$ counterexamples (rather than only one) where their labels are opposite to $g_{\theta}(\boldsymbol{x})$. If that many solutions are found, then the majority decision for the group of assignments specified by $\boldsymbol{x}$ is opposite to $g_{\theta}(\boldsymbol{x})$, otherwise, the prediction remains unchanged. The general scheme of \textit{CE-Fair Counting} approach is shown in Algorithm \ref{alg:mip_exp}. The algorithm takes as an input $f_{\theta}$, as well as a sample $\boldsymbol{x}$. In line 3, the MILP encoding of $f_{\theta}$ is obtained as per Equation \ref{eq:mip} and constraints from Equation \ref{eq:ce} specifying a counterexample for $\boldsymbol{x}$ are obtained in the following line. In line 5, the MILP search tree is explored to find %the specified number of such 
counterexamples; if it finds less than $\ceil*{\frac{|A(\boldsymbol{x})|}{2}}$, the final prediction does not change, otherwise, it flips (lines 6-9). %\golnoosh{can we say anything here wrt the runtime of MILP for verification?}

%\golnoosh{to make it clear, can we use the same terminology in definition4 and the function that is used here called: "CounterExampleEncoding"?}
%\golnoosh{we need to make sure that the definition is clear and then we can remove algorithm 1 to save space for the results}
\begin{algorithm}[t]
\caption{Counterexample-guided Counting to Guarantee Fair Predictions}
\label{alg:mip_exp}\footnotesize{
    \begin{algorithmic}
        \STATE \textbf{Input}: $f_\theta$, $\boldsymbol{x}$
        \STATE \textbf{Output}: CE-Fair Prediction: $h \left(f_{\theta}(\boldsymbol{x}) \right) \in \{0, 1\}$
        \STATE $\phi_{N} \gets {\tt ModelMIPEncoding}(f_{\theta})$ \COMMENT{Constraints in Equation \ref{eq:mip}}
        \STATE $\phi_{CE} \gets {\tt FairCEEncoding}(\phi_{N}, \boldsymbol{x}, f_{\theta}(\boldsymbol{x}))$ \COMMENT{Constraints in Equation \ref{eq:ce}}
        \STATE $\mathcal{S} \gets {\tt FairCECounting}(\phi_{N}, \phi_{CE})$ 
        \COMMENT{Constraints in Equation \ref{eq:ceset}}
        \IF{~$|\mathcal{S}| < \ceil*{\frac{|A(\boldsymbol{x})|}{2}}$}
            % \COMMENT{A(x) in Equation \ref{eq:assingment}}
            \STATE \textbf{return} ~$g_{\theta}(\boldsymbol{x})$
        \ELSE
            \STATE \textbf{return} ~$1-g_{\theta}(\boldsymbol{x})$
        \ENDIF
    \end{algorithmic}}
\end{algorithm}
%\golnoosh{what do we mean by "findSolutions" in the algorithm? can we change the name? also this should be defined in Definition 6.}

%\aish{if time permits, can we show that the inference time of enumeration is higher than the next approach maybe for one dataset like german? If not maybe we can do pen-and-paper analysis of increase in search space with these two approaches and provide a theoretical bound on why the next approach is better?}
%\golnoosh{that's a good idea, Aish! Can you give it a try to provide some theoretical analysis of the inference time?}

%\golnoosh{the following paragraph sounds very negative, I'd change it but also it would be good to present the results when comparing inference time of a naive enumeration to our approach, do we have these results}
%Although more efficient than pure enumeration, this still incurs a high runtime at inference. We can, however, relax the guaranteed fairness requirement and trade training time for inference time. In the next section, we explain how counterexamples can be leveraged in the training phase as an effective way to impose fairness.
%\begin{wrapfigure}{l}{0.3\columnwidth}

\subsection{Empirical Evaluation of CE-Fair Prediction} \label{sec:predictionresults}
This section shows the effectiveness of \textsc{CE-Fair} prediction approach through empirical evaluations on three widely known real-world benchmark datasets: German \cite{german_dataset} ($\mathcal{S} \in $ \{age, sex/marital status, foreign worker\}), IPUMS Adult \cite{newadult_dataset}($\mathcal{S} \in $ \{age, marital status, race, native country, sex\}), and Law School \cite{lawschool_dataset}($\mathcal{S} \in $ \{race, gender\}). We conduct grid-search to identify the best parameters. % for baseline model (NN\textsubscript{b}). %For each dataset, we identify the best baseline architecture and parameters by conducting grid-search and learn the best baseline model (NN\textsubscript{b}).
Experiments are implemented in Python using Pytorch~\cite{pytorch19}. %, with ADAM optimizer~\cite{kingma2014adam} to perform stochastic optimization of the neural network models. 
All experiments were run on a machine with 10 GiB RAM and a 2.1GHz Intel Xeon processor. We use Gurobi-9.5.1\footnote{\url{https://www.gurobi.com}} as our backend solver to generate counterexamples. We make all code, datasets, and preprocessing pipelines publicly available. 
%to ensure reproducibility of our results. 
We provide a more detailed overview of the datasets, experiment setup, model architectures, and hyperparameters in the Appendix.

%\aish{this paragraph needs more work, we need to write why its a good idea to choose rho. i.e. why should the reviewers buy that our results on rho will generalize to the rest of the dataset. Also make it clear that majority baseline and pretrained NN were tested on this rho percent of data}
%We highlight the fact that the search for counterexamples is an expensive process and that the run time grows with the dimensionality of the sensitive features, as well as the size of the dataset. To make our approach scalable, we introduce a hyperparameter $\rho$ that indicates what portion of the dataset we are taking. For the experiments in this section, 
%we are not concerned with $\rho$ for the train set but the portion that we take from the test set are $100\%$ for German, $10\%$ for Adult, $1\%$ for IPUMS Adult, and $100\%$ for Law School.

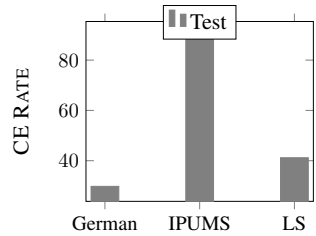
\begin{figure}[h]
\centering
%\vspace{-0.35cm}
\scalebox{0.7}{
 {\scalefont{1.1}
    \begin{tikzpicture}
    \begin{axis}[
    ybar,
    bar width=15pt,
    enlargelimits=0.10,
    legend style={at={(0.5,0.9)},
      anchor=south,legend columns=-1},
    ylabel={\large \textsc{CE Rate}},
    symbolic x coords={German, IPUMS, LS},
    xtick=data,
	width=5.9cm, height=5.0cm,     % size of the image
    ]
% \addplot+[black!90] plot coordinates {(German,7.11) (Adult,50.85) (IPUMS, 50.96) (LS, 50.96) };
\addplot+[black!50] plot coordinates {(German,29.8) (IPUMS, 89.4) (LS, 41.18) };

\legend{\strut {\large Test}, \strut {\large Train}}
\end{axis}
    \end{tikzpicture}
    }
    }
    \caption{Empirically, the best learned NN model is not fair. The figure presents the CE Rate for German, IPUMS Adult (IPUMS), and Law School (LS).
    }
    %\vspace{-0.7cm}
    \label{fig:cerate}
\end{figure}

\noindent\textbf{Q1: Does a deep neural network trained on data obey individual fairness?}
To quantify the degree of \emph{unfairness} in the initial model trained on data, we introduce, Counterexample Rate \textsc{(CE Rate)}, which computes how many test samples have counterexamples as per Definition~\ref{def:fairness_verification}. As shown in Figure~\ref{fig:cerate}, the degree of fairness violation based on these metrics is high for all our datasets, motivating the need for \emph{guaranteed} fair predictions. The percentage of data points that have counterexample can be as high as 89\% for \textsc{IPUMS Adult} dataset.

%\end{wrapfigure}
% 2) \textsc{Average Violation (Avg. Vio.)}, computed as an average ` unfairness is in the initial pretrained model
% .. 
% with an average violation of K for the train and test data. See Table~\ref{} in the appendix for detailed results on all datasets.   

% To what extent does our initial learned model violate individual fairness at prediction?}
% The first row of Table \ref{tab:german} quantifies the amount of \emph{unfairness} in the initial model trained purely to increase performance. We observe a high amount of CE Rate, as well as CE Count. These two metrics explore the amount of unfairness in the whole input space. Moreover, Flip Rate indicates that a non-negligible portion of individuals in the test set are facing an unfair outcome.
% \aish{maybe a bar plot of ce rate for test and train for each of the dataset?}

% While the mentioned metrics only indicate the frequency of fairness notion being disregarded, the average and maximum violations of the 0-th epoch in Figure \ref{fig:german} and \ref{fig:adult} indicate how severe unfairness is in the initial pretrained model. Moreover, all fairness metrics drop to zero as we have guaranteed fairness in this case.

\noindent\textbf{Q2: What is the effect of guaranteed fair predictions on performance and overall model fairness?}
In this experiment, we compare the accuracy of the best baseline model (NN\textsubscript{b}) with \textsc{CE-Fair} predictions on test data. Further, to quantify the effect of fair predictions on model fairness, we introduce \emph{flip rate} which computes the number of samples in test data where the prediction of the model was modified to satisfy individual fairness. Table~\ref{tab:ce-fair-predictions} demonstrates that you can use \textsc{CE-Fair} predictions to guarantee fairness with accuracy loss up to 8\%. This can be explained as follows: since the flip rate of the best-trained model (NN\textsubscript{b}) is high, it is expected that the drop in accuracy has a relation with the model flip rate, as seen with \textsc{German} with the lowest flip rate and smallest decrease in accuracy. Further, we compare our approach against a naive majority-based baseline which is a constant predictor that returns the most frequent label. We observe that, on average, \textsc{CE-Fair} predictions perform 8\% better than the majority baseline. In Section~\ref{sec:training}, we propose a counterexample-guided re-training approach to reduce the performance drop while improving fairness metrics. 

% \usepackage{booktabs}
\begin{table}[h]
\centering
\caption{The effect of guaranteed fair predictions on performance and model fairness -- 5-fold CV results}
\resizebox{0.9\linewidth}{!}{\normalsize{
\begin{tabular}{@{}l|ll|ll|ll@{}}
\toprule
\textbf{Dataset} & \multicolumn{2}{c|}{\textbf{NN\textsubscript{b}}}                                               & \multicolumn{2}{c|}{\textbf{Majority Baseline}}                                 & \multicolumn{2}{c}{\textbf{CE-Fair Prediction}}                                \\ \midrule
                 & \multicolumn{1}{c}{\textbf{Accuracy}} & \multicolumn{1}{c|}{\textbf{Flip Rate}} & \multicolumn{1}{c}{\textbf{Accuracy}} & \multicolumn{1}{c|}{\textbf{Flip Rate}} & \multicolumn{1}{c}{\textbf{Accuracy}} & \multicolumn{1}{c}{\textbf{Flip Rate}} \\ \midrule
German           & $76.70 \pm 2.78$                      & $8.90 \pm 0.73$                        & $70.00 \pm 1.78$                      & $0.0 \pm 0.0$                           & $74.20 \pm 3.50$                      & $0.0 \pm 0.0$                          \\
IPUMS Adult      & $81.57 \pm 0.47$                      & $24.39 \pm 1.52$                       & $54.61 \pm 0.55$                      & $0.0 \pm 0.0$                           & $73.23 \pm 0.93$                      & $0.0 \pm 0.0$                          \\
Law School       & $82.72 \pm 0.19$                                & $17.53 \pm 0.59$                                 & $72.98 \pm 0.50$                                 & $0.0 \pm 0.0$                           & $74.36 \pm 1.95$                                & $0.0 \pm 0.0$                          \\ 
\bottomrule
\end{tabular}}}
\label{tab:ce-fair-predictions}
\end{table}

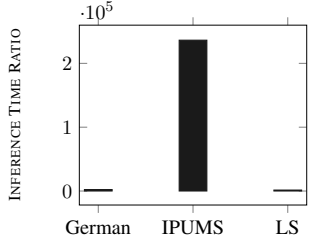
\begin{figure}[h]%{c}{\columnwidth}
\begin{center}
%\centring
%\vspace{-0.1cm}
\scalebox{0.7}{
 {\scalefont{1.1}
    \begin{tikzpicture}
    \begin{axis}[
    ybar,
    bar width=15pt,
    enlargelimits=0.10,
    legend style={at={(0.5,0.9)},
      anchor=south,legend columns=-1},
    ylabel={\small \textsc{Inference Time Ratio}},
    symbolic x coords={German, IPUMS, LS},
    xtick=data,
	width=5.9cm, height=5.0cm,     % size of the image
    ]
\addplot+[black!90] plot coordinates {(German,2368) (IPUMS, 236351) (LS, 1560) };
% \legend{\strut {\large Train}, \strut {\large Test}}
\end{axis}
    \end{tikzpicture}
    }
    }
    \caption{Ratio of Inference time increases when using CE-Fair predictions to guarantee fairness. %This is expected due to the high flip rate in the best trained model (see Table~\ref{tab:ce-fair-predictions}).
    }
    %\vspace{-0.6cm}
    \label{fig:ceinftime}
    \end{center}
\end{figure}

% As the first two rows in Table \ref{tab:german} indicate, we would not lose accuracy substantially, however, the extra per-sample run-time incurred at inference is high when we want to guarantee fairness.
\noindent\textbf{Q3: How does \textsc{CE-Fair} prediction affect inference time?}
Figure~\ref{fig:ceinftime} plots the ratio of inference time of NN\textsubscript{b} and \textsc{CE-Fair} predictions for test data on all datasets. We observe that the increase in inference time is proportional to the model flip rate (see Table~\ref{tab:ce-fair-predictions}). It also highly depends on the dimension of the sensitive features as seen with \textsc{IPUMS Adult}. This is expected since a larger flip rate means more samples are unfair and therefore more calls to the verification engine. Of course, when violating fairness leads to ethical or legal problems, the question is not whether we can afford fairness enforcement, but whether it is correct to use machine learning at all. In this context, the computational price of enforcing fairness, even if it ends up being significant, is entirely warranted. 

%\section{Counterexample-Guided Fairness Enforced Training~(CE-Fair Training)}
\section{CE-Fair Training}
\label{sec:training}

\begin{figure*}[ht]
    \begin{subfigure}{0.32\textwidth}\includegraphics[width=\textwidth]{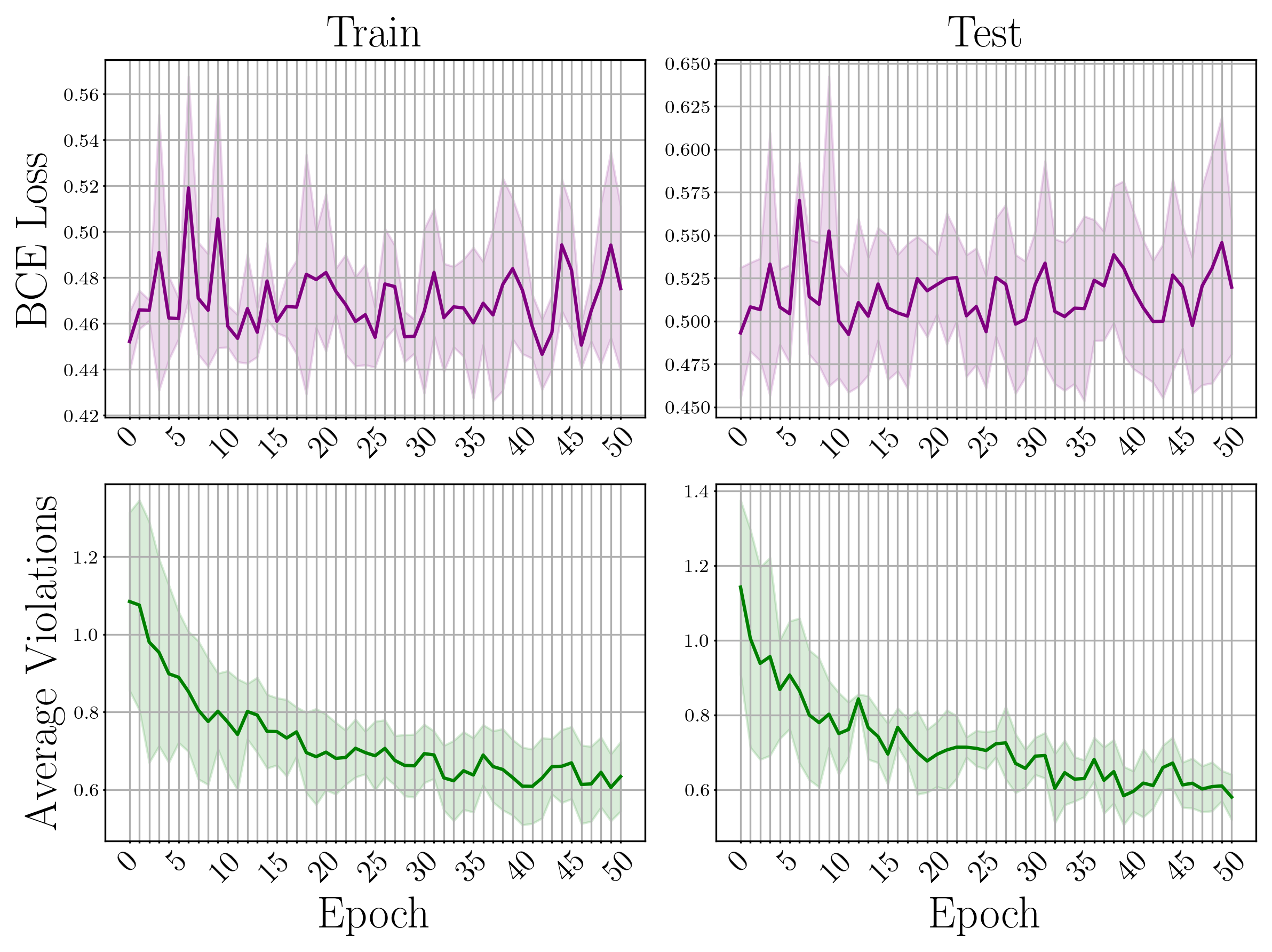}
        \caption{German (\textsc{full batch}).}
        \label{fig:training:german}
    \end{subfigure}
    \hfill
    \begin{subfigure}{0.32\textwidth}\includegraphics[width=\textwidth]{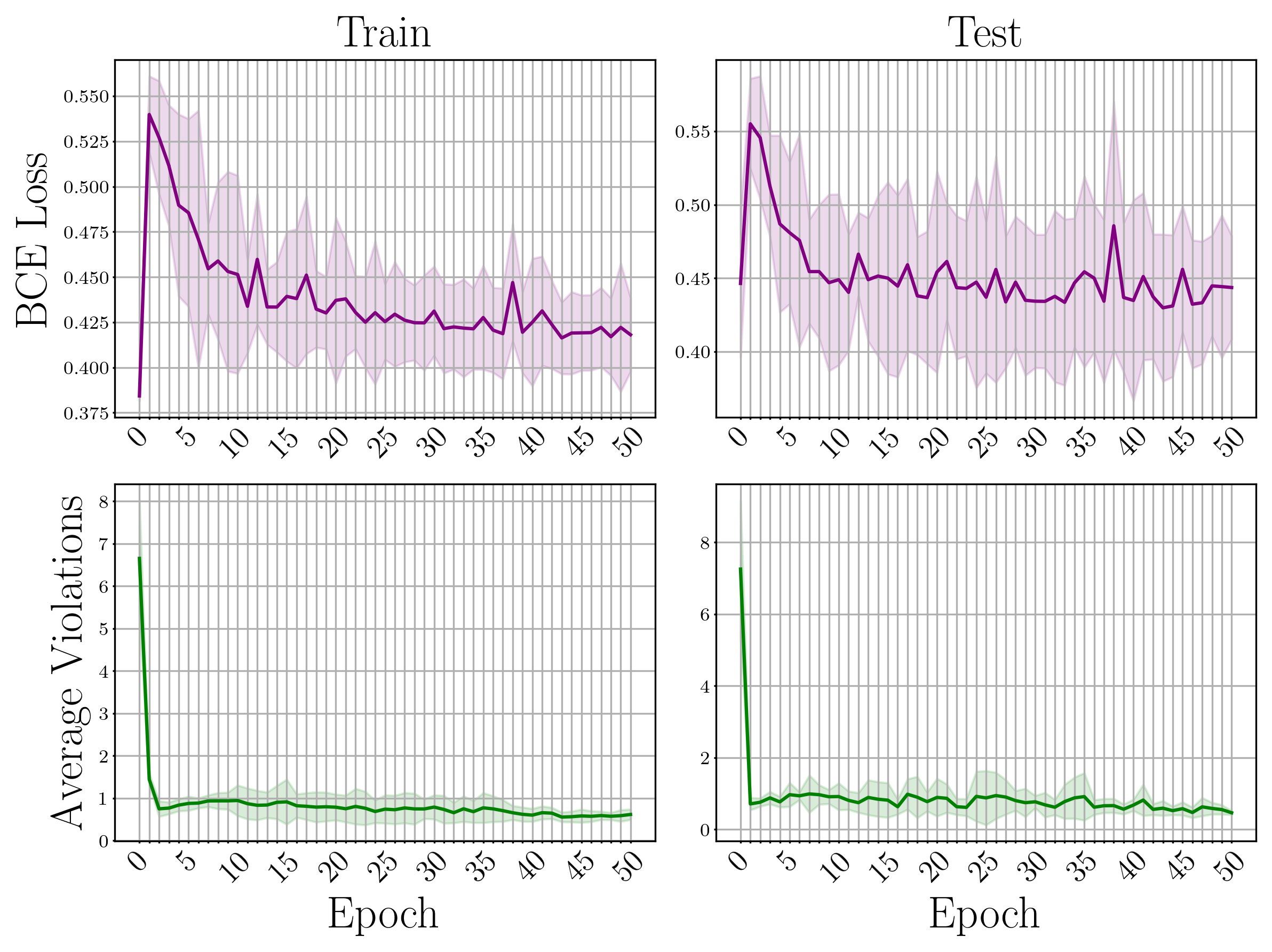}
        \caption{IPUMS Adult (\textsc{CE batch}).}
        \label{fig:training:newadult}
    \end{subfigure}
    \hfill
    \begin{subfigure}{0.32\textwidth}\includegraphics[width=\textwidth]{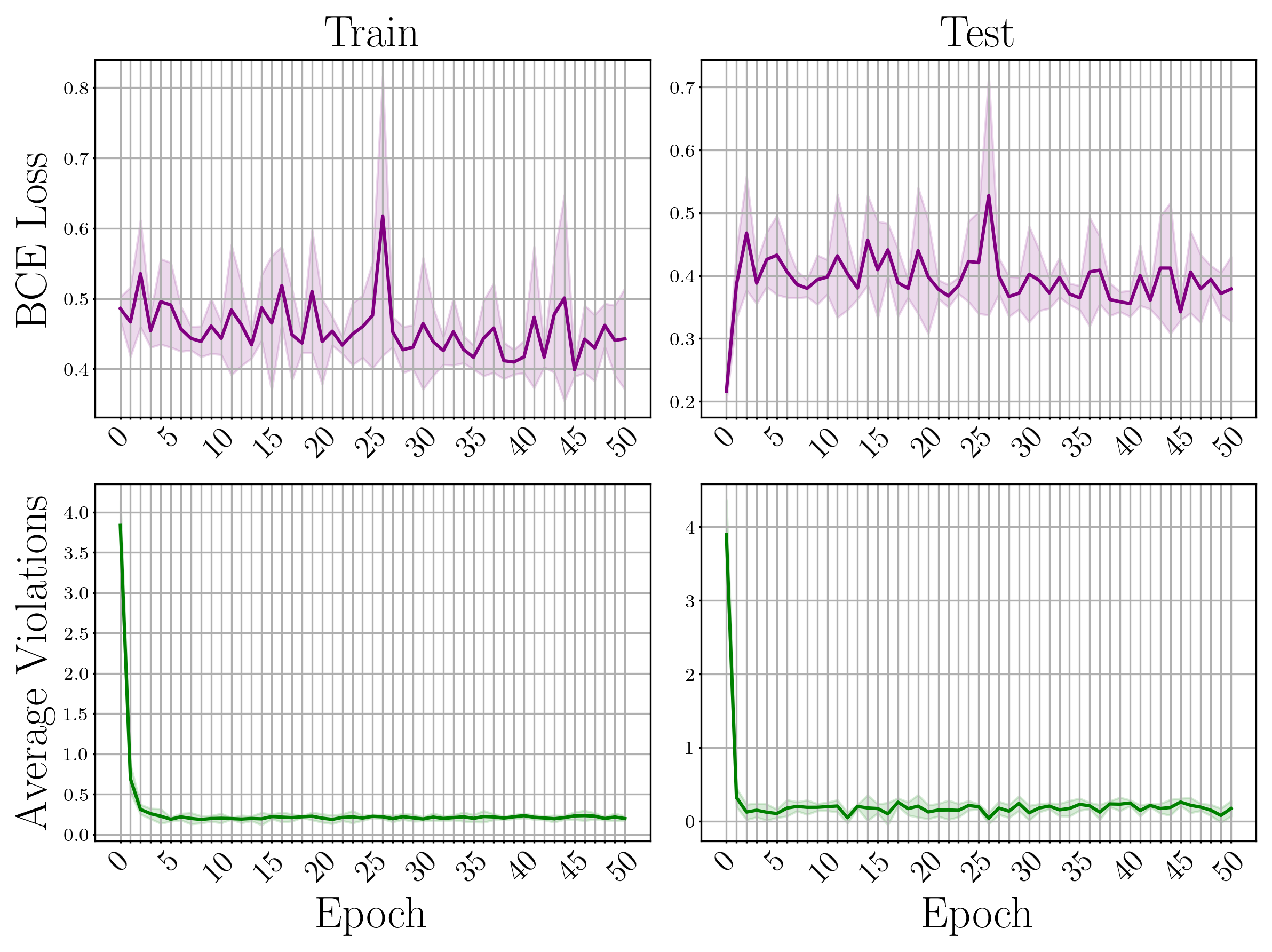}
        \caption{Law School (\textsc{CE batch}).}
        \label{fig:training:lawschool}
    \end{subfigure}
    \caption{Loss and average violation curves for fairness-enforced training on different datasets. Each curve is the mean over 5 runs and the shaded area represents the std.}
    \label{fig:training}
\end{figure*}

%\golnoosh{define fair learning based on the definition of individual fairness introduced in the methodology section}
%\golnoosh{Let's first define Individual fairness verification that return CEs and then we can explain how we use the verifier to guarantee fair prediction and then introduce  counterexample-guided fair prediction algorithm (CE-Fair prediction)}
%\golnoosh{the text of this section needs re-write! and fairness training should be defined formally not verbally. }
%\golnoosh{Add a paragraph to motivate the need to re-training, and then define NN main objective function and then add the CE inductive bias to the loss function}
%\golnoosh{this paragraph needs a rewrite}
In this section, we propose an algorithm for learning fair neural networks with counterexamples. %Our learning algorithm works by training a typical network with a counterexample-driven fairness regularization. 
While in the previous section, we guaranteed fair predictions as a \emph{post-processing} approach, in this section, we propose an \emph{in-processing} approach that drops the guaranteed fairness requirement in exchange for a relatively fair model with efficient inference. Our learning algorithm is orthogonal to the prediction technique of the previous section and both approaches can be combined to acquire guaranteed fairness with boosted performance and more efficient inference time (see evaluation results in Section~\ref{sec:trainingresults}). 

%Fairness constraints are extra requirements in the whole process of training and deploying a model. These extra requirements incur extra effort, in this case, in the form of a trade-off among training time and inference time. If fast inference is not crucial, we apply the method in Section \ref{sec:majority} and we get guaranteed fair predictions. In this section, we take the case that inference needs to be efficient. Assume that a stake-holder has provided us with a model, pre-trained with no fairness constraints. We would like to assess whether this model is fair, and if not, fine-tune it towards being more fair. We spend more training time and we drop the guaranteed fairness requirement in exchange for a relatively fair model with efficient inference. Of course, we can still apply the method in section \ref{sec:majority} on top to acquire guaranteed fairness (see evaluation results in Section~\ref{}). \kiarash{elaborate based on results.}

To define our learning paradigm, we first define a specific variation of the verification function defined in Definition~\ref{def:fairness_verification} called $v_{max}(f_{\theta}, x)$. The \textsc{CE-Fair} verification function can produce a counterexample given an arbitrary sample. To make sure that the counterexamples are representative and not out-of-distribution examples for training, we generate counterexamples relative to each training point. Moreover, to make sure that the counterexamples are effective in reducing the degree of fairness violation, we instead appeal to Definition~\ref{def:maxverification} to generate counterexamples with maximal violation relative to each training point. 
%such that given a data point $x$ and a trained NN $f_{\theta}$ finds the CE data point that has the maximum violation.
\begin{definition} \label{def:maxverification}(Maximum Violation \textsc{CE-Fair} Verification) Given a data point $\bm{x}$ and $f_{\theta} :\mathcal{X} \rightarrow [0, 1]$, we find the maximum violation counterexample $\bm{x}_{max}^\prime$ using $v_{max}(f_{\theta}, \bm{x})$ defined as:
\end{definition}
{\footnotesize\begin{equation}
\label{eq:maxviolationCE}
     \argmax_{\bm{x}^\prime}~ |f_{\theta}(\bm{x}^\prime) - f_{\theta}(\bm{x})| \quad s.t. \quad x^\prime = v(f_{\theta}, \bm{x})
\end{equation}}

%\golnoosh{why in the algorithm for y=0 and y=1 we have different lines?}
%\kia{Kiarash: I was trying to contrast the two cases: For example, when $y=0$ it means that the model had negative output logit and we should maximize its output for the CE (to be as positive as possible for maximum violation!). But if it is confusing, we could as well merge the two cases into one line: $CE,~ vio \gets {\tt Optimize}(\phi_{N}, \phi_{CE})$}
%\golnoosh{You can merge it and call the function the same as the definition 8 (not optimize which is not clear what do you mean by it).}
Suppose a sample data point $\bm{x}$, is a young single Asian female applicant with a negative decision on her loan application in the training set. Using $v_{max}(f_{\theta}, \bm{x})$, we are able to find a counterfactual applicant $\bm{x}^\prime$ with a different combination of sensitive attributes, e.g., an old married American man with the same credit history who can potentially receive a positive decision from the model with the highest level of violation compared to the sample data $\bm{x}$. Next, we show how we use such counterfactual examples, i.e., counterexamples, to train a fair model.
%\golnoosh{these two definitions should be connected.}

%\golnoosh{this definition needs to be completed: and flip of the label should be added to the fairness loss}
\begin{definition} (\textsc{CE-Fair} Training) The loss function for counterexample-guided training can be written as:

{\footnotesize \begin{equation}
\begin{split}
& L_{CE-Fair}(f_\theta) = \underbrace{{L_{BCE}(f_\theta)}}_{\textnormal{Classification loss}} +\\ 
&  \underbrace{\sum_{i=1}^{M} y_i \log(f_\theta(v_{max}(f_\theta, \bm{x}_i)))) + (1-y_i) \log(1 - f_\theta(v_{max}(f_\theta, \bm{x}_i))))}_{\textnormal{Fairness Counterexample-guided loss}}
\end{split}
\end{equation}}
where $L_{BCE}$ is the classification loss defined in Equation~\ref{eq:loss} and $M$ denotes samples in the training dataset. Note that $v_{max}(f_{\theta}, x_i)$ is obtained for each datapoint $x_i$ by solving a MILP program with an objective that is defined in Equation~\ref{eq:vio_obj} and all the constraints in Equations~\ref{eq:mip} and \ref{eq:ce}.
\end{definition}

%\subsection{Counterexample-guided Fairness Training (Fair-CE Training)}

%\golnoosh{this needs to be written with reference to the algorithm similar to the previous algorithm using lines, etc.}
Algorithm \ref{alg:feta} summarizes the counterexample-guided training. %loop 
%to fine-tune a pretrained model toward fairness with batch-training. At each batch, for each datapoint $\bm{x}$, we search for a counterexample $\bm{x}^\prime$ with maximum \emph{violation}, i.e., a counterexample that would not only flip the prediction but also maximally changes the logit output of the model w.r.t. this datapoint. We then include this counterexample with the correct \emph{fair} label (opposite to what the model currently returns for it) in the train batch. As it is an extreme case of a counterexample with maximum violation, we effectively regularize the model toward being fair by updating the model with the new batch. \kiarash{elaborate the effectiveness of the regularizer as it fine-tunes the model already in the first few epochs.}
Note that our CE-Fair training incorporates data augmentation through counterexamples which can cause drift in the model quality. Our approach guards against this by recomputing counterexamples for each batch at every epoch (i.e., lines 7-9). This ensures that: i) an incorrect old counterexample does not burden the learning, and ii) learning incorporates multiple counterexamples from the training set at a time and so is less sensitive to any particular one. There are different heuristics that one could adopt to use counterexamples and encourage the learned function to
become fairer. In Line 11, we allow the user to keep all the original samples in the batch (aka, \emph{full batch}) to preserve accuracy. Contrary to this, we can only keep the original samples for which we have found a counterexample (aka, \emph{CE batch}). The latter is effective in reducing discrimination, specifically when the counterexample-guided training is initiated with an optimal $f_{\theta}$. This introduces a tradeoff as to what portion of the original samples to keep. In our empirical evaluation, we only evaluate based on \emph{full batch} and \emph{CE batch} for simplification.

%\kia{Consider adding this explanation here instead of in Q6: "Moreover, as can be seen in line 11, we are keeping all the original samples in the batch (aka, \emph{full batch}) to preserve accuracy. Contrary to this, we can only keep the original samples for which we have found a counterexample (aka, \emph{CE batch}). This introduces a tradeoff as to what portion of the original samples to keep. However, we only try \emph{full batch} and \emph{CE batch} for simplification."}

%\golnoosh{This algorithm can change and we can show the gradient update instead}

\begin{algorithm}[h]
\caption{Counterexample-guided Fairness Enforced Training}
\label{alg:feta}{\footnotesize
    \begin{algorithmic}
        \STATE \textbf{Input}: $f_{\theta}$, $\bm{D}$, $\rho$
        \STATE \textbf{Output}: CE-Fair Fine-tuned model: $f_{\theta}$
        \FOR{$epoch \in \{1 \dots e\}$}
            \FORALL{$batch \in {\tt shuffled}(\bm{D})$}
                \STATE $sampled\_batch \gets {\tt RandomSample}(batch,\rho)$
                \FORALL{$(\bm{x}, y) \in sampled\_batch$}
                    \STATE $\phi_{N} \gets {\tt ModelMIPEncoding}(f_{\theta})$ \COMMENT{Constraints in Equation \ref{eq:mip}}
                    \STATE $\phi_{CE} \gets {\tt FairCEEncoding}(\phi_{N}, \boldsymbol{x}, f_{\theta}(\boldsymbol{x}))$ \COMMENT{Constraints in Equation \ref{eq:ce}}
                    \STATE $x_{max}^\prime \gets {\tt FindMaxViolationCE}(\phi_{N}, \phi_{CE})$ \COMMENT{Objective as in Equation~\ref{eq:maxviolationCE}}
                    \IF{$x_{max}^\prime$ exists}
                        \STATE 
                        %$sampled\_batch \gets 
                        {\tt Append}$(sampled\_batch, (x_{max}^{\prime}, y))$
                    \ENDIF
                \ENDFOR
            \STATE $\theta \gets {\tt OptimizationStep}(f_{\theta}, sampled\_batch)$
            \ENDFOR
        \ENDFOR
\end{algorithmic}}
\end{algorithm}
%Concretely, given the MIP encoding \ref{eq:mip}, we would like the counterexample $\bm{x}^\prime$ that satisfies:

%\begin{equation}
%    \max_{\bm{x}^\prime}~ |\bm{z}_n - f_{\bm{x}}|
%\end{equation}

%where $f_{\bm{x}}$ is the output logit of the NN for $\bm{x}$ and $\bm{z}_n$ is the pre-ReLU value of the variable corresponding to the output of the network.

%\golnoosh{This is not quite readable. Instead of explaining it in the text, please define all the notions like the previous section, starting with the pre-trained model and then you can define the verification function that gets a trained NNs as a data point as an input and return a CEs that satisfy the above equation. Using that function, you can then write the actual objective where you add the regularizer that use the verification function to return a counterexample with maximum violation. }

%\golnoosh{When you add an algorithm, you need to explain all the steps to the reader in the main body of the text. You need to first define all the notion precisely and then you can use the same one in the algorithm.}

\textbf{Practical Considerations.} We highlight the fact that the search for counterexamples is an expensive process and the run time grows with the size of the $D$, the number of epochs $e$, and the dimensionality of sensitive features. To make our approach scalable, we introduce a hyperparameter $\rho$ that indicates what portion of the dataset we are taking. In Section~\ref{sec:trainingresults}, we show that even a small value of $\rho$, as small as $1\%$, is effective to fine-tune the model to become fairer in only a few epochs.

%We might not be able to leverage the complete dataset for fairness-enforced retraining even though the initial model had been pretrained on the whole dataset. For this reason, we introduce a ratio parameter $\rho$ that indicates what portion of the dataset we are taking. Moreover, we might not be able to retrain for large number of epochs; in fact, we will show in Section~\ref{sec:eval} that these maximum violated counterexamples are effective enough to fine-tune the model in only a few epochs.

%\aish{bring back the rho thing here, i think we have a paragraph earlier.}

\textbf{Multi-objective Model Selection.} In fairness-enforced re-training, we are concerned with accuracy and unfairness at the same time. We thus opt for choosing a \emph{Pareto frontier} by selecting the epoch whose accuracy and unfairness, when seen as a point in the $2D$ space, have the minimum $\ell_2$ distance to the point corresponding to maximum accuracy and minimum unfairness. 

\textbf{FETA Extension.} %Similar to CE-Fair Verification in Section~\ref{sec:verification}, e
Extending \textsc{CE-Fair} prediction and training to multi-class classification (one-vs-rest approach) is straightforward. To extend these approaches to regression, e.g., to predict credit limit, we need to modify Equation~\ref{eq:ce_b} to encode fairness counterexamples as $|f_{\theta}(\bm{x}) - f_{\theta}(\bm{x}^\prime)|>\epsilon$ where $\epsilon$ is a hyperparameter that needs to be defined based on the context. Also, we need to consider an appropriate loss function, e.g., $\ell_2$ loss for regression, to include counterexamples in the training. Note that our neural network MILP encoding is not limited to ReLU and can encode any piece-wise linear activation function. 
%\golnoosh{Why you return vio when vio is not used anywhere?}
%\golnoosh{You need to specify the groud truth label for CE when adding it to the sampled batch}

%\golnoosh{motivate the need for training a fair model and connect this section to the prediction section where the cost of inference is high (it is interesting to add a table in the paper and present these differences in terms of cost of training and time of inference) at the end of this section after introducing all three variations. We can then refer the reader to the empirical evaluation of the methods in section 6.}

\subsection{Empirical Evaluation of \textsc{CE-Fair} Training}\label{sec:trainingresults}
In this section, we evaluate the learning algorithm both on its own and in conjunction with the prediction technique. We use the same datasets and hardware as in Section~\ref{sec:predictionresults}. 

\textbf{Q4: Does our \textsc{CE-Fair} training algorithm make the original unfair model fairer?} In this experiment, we re-train NN\textsubscript{b} model with $\rho = 100\%, 1\%, 2\%$ for German, IPUMS Adult, and Law School, respectively. We train for 50 epochs and select the best model based on the \emph{Pareto frontier} discussed in Section~\ref{sec:training}. Figure \ref{fig:training} summarizes the learning curves. We observe that while loss oscillates due to the fairness-performance tradeoff, the average of maximum violation substantially decreases. The results presented here are chosen among the \textsc{full batch} and \textsc{CE batch} options. \textsc{CE batch} decreases average violation almost to its minimum only in the first few epochs. This is because it is focusing only on the counterexamples while \textsc{full batch} experiences a more smooth curve.

\begin{table}[h]
\centering
\caption{The effect of \textsc{CE-Fair} re-training on fairness metrics -- 5-fold CV results}
\resizebox{\linewidth}{!}{\normalsize{
\begin{tabular}{l|l|lll}
\hline
%\textbf{Dataset} & \multicolumn{3}{c|}{\textbf{NN\textsubscript{b}}}                                                                                      & \multicolumn{3}{c}{\textbf{CE-Fair Training}}                                                                         \\ \hline
                \textbf{Dataset}
                 & \textbf{Approach} &\multicolumn{1}{c}{\textbf{Accuracy}} & \multicolumn{1}{c}{\textbf{Flip Rate}} & \multicolumn{1}{c}{\textbf{CE Rate}}  \\ \toprule
German          & {NN\textsubscript{b}} & $76.70 \pm 2.78$                      & $8.90 \pm 0.73$                        & $29.80 \pm 2.15$                      \\
&{CE-Fair Training}& \textbf{$75.70 \pm 3.77$ }                     & \textbf{$3.40 \pm 0.73$}                         & \textbf{$17.70 \pm 4.67$}                     \\\midrule

IPUMS Adult      & {NN\textsubscript{b}} & $81.57 \pm 0.47$                      & $24.39 \pm 1.52$                       & $89.40 \pm 1.80$                    \\  &{CE-Fair Training}& \textbf{$80.03 \pm 0.57$}                     & \textbf{$2.80 \pm 0.67$}                        & \textbf{$15.34 \pm 2.53$}                     \\\midrule
Law School       & {NN\textsubscript{b}} & $82.72 \pm 0.19$                      & $17.53 \pm 0.59$                       & $41.18 \pm 2.05$                      \\&{CE-Fair Training}& \textbf{$84.99 \pm 0.28$}                      & \textbf{$5.06 \pm 0.97$}                        & \textbf{$7.64 \pm 0.89$ }                     \\ \bottomrule
\end{tabular}}}
\label{tab:ce-fair-training}
\end{table}

To quantify if the function is fairer, we compare two fairness metrics \textsc{Flip Rate} and \textsc{CE Rate} defined in Section~\ref{sec:predictionresults}. As shown in Table~\ref{tab:ce-fair-training}, counterexample-guided retraining leads to better fairness metrics on all datasets by reducing the number of fairness violations. In fact, in the IPUMS Adult dataset, we see the highest decrease of 74\% w.r.t. \textsc{CE Rate}. These results indicate the usefulness of \textsc{CE-Fair} learning to make the original unfair model fairer. Further, the drop in accuracy when enforcing fairness is negligible when compared to the original model; In fact, we observe an increase in accuracy for Law School. While \textsc{CE-Fair} training significantly reduces fairness violations, it does not guarantee fair predictions for all points in the input domain. This motivates the need for using \textsc{CE-Fair} predictions in conjunction with the counterexample-guided learning algorithm, to guarantee fair predictions.

\begin{table}[h]
\centering
\caption{Comparison of applying \textsc{CE-Fair} Prediction on \textbf{NN\textsubscript{b}} vs. on the \textsc{CE-Fair} re-trained model -- 5-fold CV results}
\resizebox{\linewidth}{!}{\normalsize{
\begin{tabular}{l| l | cc}
\hline
\textbf{Dataset} %& \multicolumn{2}{c|}{\textbf{CE-Fair Prediction}}                                     & \multicolumn{2}{c}{\textbf{CE-Fair Training+ CE-Fair Prediction}}                                                   \\ \hline
                & \textbf{Approach} & \multicolumn{1}{c}{\textbf{Accuracy}} & \multicolumn{1}{c}{\textbf{Inference Time (s)}} %& \multicolumn{1}{c}{\textbf{Accuracy}} & \multicolumn{1}{c}{\textbf{Inference Time (s)}} 
                \\ \toprule
German          & CE-Fair Prediction & $74.20 \pm 3.50$                      & $0.45 \pm 0.06$     \\                   & CE-Fair Training+ CE-Fair Prediction & $75.30 \pm 3.65$                      & $0.39 \pm 0.03$                             \\\midrule

IPUMS Adult       & CE-Fair Prediction & $73.23 \pm 0.93$                      & $119.64 \pm 33.45$                         \\                   & CE-Fair Training+ CE-Fair Prediction  & $79.46 \pm 0.71$                      & $10.20 \pm 2.90$                              \\\midrule
Law School       & CE-Fair Prediction & $74.36 \pm 1.95$                      & $0.39 \pm 0.14$     \\                   & CE-Fair Training+ CE-Fair Prediction                         & $84.14 \pm 0.97$                      & $0.30 \pm 0.13 $                            \\ \bottomrule
\end{tabular}}}
\label{tab:fetafull}
\end{table}

\textbf{Q5: Does counterexample-guided learning improve the quality of the guaranteed prediction model?}
As shown in Table~\ref{tab:fetafull}, by additionally enforcing fairness constraints through counterexample-guided re-training, we improve both accuracy and inference time of \textsc{CE-Fair} predictions. Running \textsc{CE-Fair} predictions on the re-trained model improves inference runtime significantly on IPUMS Adult which has the largest sensitive feature space. The maximum drop in accuracy compared to NN\textsubscript{b} is  only 1.5\%. Whereas, running \textsc{CE-Fair} predictions directly on NN\textsubscript{b} leads to a maximum accuracy loss of 8.3\%. Thus, with \textsc{CE-Fair} training, we get both a fairness guarantee and better runtime and model performance.  
%\aish{call out Law school dataset and explain what kiarash mentioned in slack: with just re-training on 2\% of the train dataset, we generalized well, reduced number of CEs and lead to better performance}

\textbf{Q6: How does {\feta} perform compared to fairness under unawareness model?}

%straight-forward approaches that satisfy the fairness criteria?}
%As shown before, a naive approach to satisfy fairness definition \ref{def:fairness} is to output a constant decision for all inputs. This constant output can be the most frequent ground-truth label in the train set, aka, the majority baseline. Another approach is to
We train a model that is \emph{unaware} of the sensitive features. Such a model would satisfy fairness definition \ref{def:fairness} as its decision is not prone to changes in the sensitive attributes. We call this model the \emph{blind} model. %Thus, once a stake-holder has a trained model that exhibits fairness violations, they face 3 ways to make it provably fair across the whole input space: i) throw it away and pick the naive majority model, ii) throw it away and train a blind model, iii) apply \textsc{CE-Fair} Predictions (\ref{sec:majority}) on top of it, iv) re-train using \textsc{CE-Fair} Training (\ref{sec:training}) and then apply \textsc{CE-Fair} Predictions (\ref{sec:majority}).
% \usepackage{booktabs}
\iffalse
\begin{table}[t]
\centering
\caption{Accuracy of different fair models compared -- 5-fold CV}
\resizebox{\linewidth}{!}{\normalsize{
\begin{tabular}{@{}l|c|c|c@{}}
\toprule
\textbf{Dataset} & 
\textbf{Majority Baseline} & \textbf{Blind Model} & \textbf{CE-Fair Training + Prediction}                                \\ \midrule
German & $70.00 \pm 1.78$ & $70.30 \pm 2.20$ & $75.30 \pm 3.65$ \\
IPUMS Adult & $54.61 \pm 0.55$ & $78.90 \pm 0.15$ & $79.46 \pm 0.71$ \\
Law School & $72.98 \pm 0.50$ & $74.19 \pm 1.20$ & $84.14 \pm 0.97$\\
\bottomrule
\end{tabular}}}
\label{tab:ce-fair-blind}
\end{table}
\fi

\begin{table}[h]
\centering
\caption{Accuracy of different fair models compared -- 5-fold CV}
\resizebox{\linewidth}{!}{\normalsize{
\begin{tabular}{@{}l|c|c@{}}
\toprule
\textbf{Dataset} & 
 \textbf{Blind Model} & \textbf{CE-Fair Training + Prediction}                                \\ \midrule
German & %$70.00 \pm 1.78$ & 
$70.30 \pm 2.20$ & $75.30 \pm 3.65$ \\
IPUMS Adult & %$54.61 \pm 0.55$ & 
$78.90 \pm 0.15$ & 
$79.46 \pm 0.71$ \\
Law School & %$72.98 \pm 0.50$ & 
$74.19 \pm 1.20$ & $84.14 \pm 0.97$\\
\bottomrule
\end{tabular}}}
\label{tab:ce-fair-blind}
\end{table}

In Table \ref{tab:ce-fair-blind}, we compare the %majority,
blind model with the %\textsc{CE-Fair} Training, and 
\textsc{CE-Fair} Training + Prediction models. We train the blind model for the same amount of epochs and the same $\rho$ (ratio) as \textsc{CE-Fair} Training for a fair comparison. Note that the blind model is fair by design but similar to \textsc{CE-Fair} Training, it requires re-training the model. We observe that \textsc{CE-Fair} Training + Prediction, aka, the FETA pipeline,  gives consistently better accuracy, up to $10\%$ better compared to the blind model, while providing similar fairness guarantees.

\textbf{Q7: How does {\feta} compare against existing work?}

The literature on fairness in machine learning contains several well-established notions, particularly group fairness notions like demographic parity, equalized odds, and equal opportunity~\cite{hardt2016equality}. However, as previously demonstrated~\cite{binns2020apparent}, it is not straightforward to achieve both group fairness and individual fairness in a single model. To compare our framework with group fairness mitigation techniques, an extension to group fairness notions would be required, but that falls outside the scope of our current work.
Table~\ref{tab:feta_lcifr} reports the
accuracy and CE Rate of {\feta} compared to a recent method called \textit{LCIFR} \cite{ruoss2020learning} that mitigates individual fairness using the same fairness definition as ours
(i.e., Definition~\ref{def:fairness}). To the best of our knowledge, \textit{LCIFR} is the closest related work to ours that includes the same fairness notion. In \textit{LCIFR} \cite{ruoss2020learning}, the authors propose a fair representation learning approach and adversarial classification to address individual fairness. We fine-tune both methods on all datasets and extend \textit{LCIFR} to accept multiple sensitive attributes similar to our setting. To gather the results of this section, we extend the sensitive features of \textit{LCIFR} and use our own train/test split, we also report the empirical results using the original implementation in the Appendix.% using the specified sensitive feature per dataset. 
We set $\gamma = 1.0$ (their loss balancing factor) for the sake of fair comparison. The results in Table~\ref{tab:feta_lcifr} indicate that the accuracy of {\feta} outperforms \textit{LCIFR} on German and Law School, and only for IPUMS Adult, \textit{LCIFR} trains a more accurate model. This finding is noteworthy because \textit{LCIFR} combines both pre-processing and in-processing techniques to enhance performance while our approach focuses on in-processing and post-processing techniques. It is worth saying that our approach guarantees to have no CE rate for all three datasets while \textit{LCIFR} cannot provide any guarantees, i.e., \textit{LCIFR} results are empirical and even if they exhibit close to zero unfairness on the test set, it does not imply guaranteed fairness over all the data points in the input space. Also, we use a combination of categorical and continious sensitive features while for \textit{LCIFR} we only use the categorical ones due to their design. Note that using continuous sensitive features results in a larger counterexample space. %\golnoosh{To gather the results of this section, we extend \textit{LCIFR} to work with multiple sensitive features, we also report the empirical results using the original implementation.% using the specified sensitive feature per dataset. The results can be found in the Appendix. %When comparing the use of one sensitive feature, our method significantly outperforms \textit{LCIFR} in both reducing the CE rate and increasing model accuracy.}

\begin{table}[h]
        \centering
        \caption{Comparison of FETA and LCIFR %\cite{ruoss2020learning}.
    }
        \resizebox{\linewidth}{!}{\footnotesize{
        \begin{tabular}{l|l| c c}
        \toprule
        \textbf{Dataset} & \textbf{Approach} &  \textbf{Accuracy (\%)} &\textbf{CE Rate (\%)} \\
        \toprule
        German & NN\textsubscript{b} & $76.70 \pm 2.78$ & $29.80 \pm 2.15$\\
        &LCIFR \cite{ruoss2020learning} & $ 72.30 \pm 1.43$ & $0.20 \pm 0.24$\\
        &\feta (CE-Fair Train. + Pred.) & $75.30 \pm 3.65$ & $0.0 \pm 0.0$\\
        \midrule
        IPUMS Adult & NN\textsubscript{b} & $81.57 \pm 0.47$ & $89.40 \pm 1.80$\\
        & LCIFR \cite{ruoss2020learning} & $81.52 \pm 0.34$ & $0.05 \pm 0.04$\\
        & \feta (CE-Fair Train. + Pred.) & $79.46 \pm 0.71$ & $0.0 \pm 0.0$\\
        \midrule
        Law School & NN\textsubscript{b} & $82.72 \pm 0.19$ & $41.18 \pm 2.05$\\
        & LCIFR \cite{ruoss2020learning} & $74.13 \pm 0.76$ & $0.008 \pm 0.007 $\\
        & \feta (CE-Fair Train. + Pred.) & $84.14 \pm 0.97$ & $0.0 \pm 0.0$ \\
        \bottomrule
        \end{tabular}}}
    
    \label{tab:feta_lcifr}
\end{table}

\vspace{-15pt}
\section{Conclusion \& Future Directions}
%\aish{i renamed this to summary and discussion, to leave at a more positive note}
\label{sec:conclusion}
In this work, we propose 1) a counterexample-guided fairness verification, 2) a counterexample-guided approach to guarantee fairness as a post-processing approach without intervening in the model%, but at the price of higher inference time
, 3) a counterexample-guided approach to adapt an already trained model toward being fair, which cannot guarantee fairness but, instead, provides fast inference, An open-source tool called {\feta} that facilitates the integration of multiple techniques for optimal results. % with an extensive evaluation on real-world datasetss. %The latter, while substantially decreasing unfairness, cannot guarantee fairness and, instead, provides fast inference. 
We showed using real-world datasets that in practice we can have efficient and fair models with little damage to accuracy. While the results of our approaches are promising, we note that the causal discrimination fairness notion adopted in this work is limited, for example, it does not capture the relationships among features. An interesting future work would be to extend {\feta} with other fairness notions. A notable example is counterfactual fairness~\cite{kusner2017counterfactual} which is a causal fairness measure based on SCM~(structural causal model). Another future avenue to explore is to bind the counterexamples to follow the distribution of the data %. One can extend our framework 
by adding distribution constraints to our MILP formulation.% to restrict counterexamples to be Out-of-Distribution. 

%%%%%%%%%%%%%%%%%%%%%%%%%%%%%%%%%%%%%%%%%%%%%%%%%%%%%%%%%%%%%%%%
% Acknowledgment 
%%%%%%%%%%%%%%%%%%%%%%%%%%%%%%%%%%%%%%%%%%%%%%%%%%%%%%%%%%%%%%%%
%\begin{ack}
%Funding support for project activities has been partially provided by Canada CIFAR AI Chair, Google scholar award, and NSERC Discovery Grants program.
%\end{ack}
\newpage
\bibliography{main}
\bibliographystyle{icml2023}

%%%%%%%%%%%%%%%%%%%%%%%%%%%%%%%%%%%%%%%%%%%%%%%%%%%%%%%%%%%%%%%%%%%%%%%%%%%%%%%
%%%%%%%%%%%%%%%%%%%%%%%%%%%%%%%%%%%%%%%%%%%%%%%%%%%%%%%%%%%%%%%%%%%%%%%%%%%%%%%

\appendix

\clearpage

\section{Datasets} \label{sec:data}

In this section, we overview the datasets used for the experiments and point out some related details.

\begin{itemize}
    \item \textbf{German credit dataset} \cite{german_dataset}\\
    This dataset consists of 1k samples with dimensionality 61 and sensitive features: age $\in [19, 75]$, sex/marital status with 4 categories, and foreign worker with 2 categories. The main task is binary classification of good or bad credit risks.
    
    \item \textbf{IPUMS Adult dataset (aka, \emph{the new Adult})} \cite{newadult_dataset}\\
    The initial Adult dataset \cite{adult_dataset} is used for binary classification of whether an individual's salary is above or below \$50k. \cite{newadult_dataset} discuss some limitations of this dataset and propose a reconstruction of the Adult \cite{adult_dataset} dataset in which the actual \emph{income} of the individuals are available. Thus, one can re-define the binary classification task with some threshold other than \$50k. In our experiments, this threshold is set to \$30k as the experiments by \cite{newadult_dataset} indicate the most severe \emph{unfairness} to occur around the 30k threshold.
    
    The dataset consists of 49k samples with dimensionality 103 and sensitive features: age $\in [17, 90]$, marital status with 7 categories, race with 5 categories, native country with 41 categories, and sex with 2 categories. This is the largest sensitive feature space among datasets used for our experiments.
    
    \item \textbf{Law School dataset} \cite{lawschool_dataset}\\
    This dataset, consisting of 86k samples, gathers law school admission records and is used for predicting if an individual would pass the bar exam. The input dimension is 37 and the sensitive features are: race with 3 categories and gender with 2 categories
    
\end{itemize}

\section{Experimental Setup and Details}

In this section, we review the experimental setup in more detail.

\begin{itemize}
    \item \textbf{Data}\\
    The datasets were discussed in Section \ref{sec:data}. The data is divided into 5 folds of 80/20 train/test sets and experiments are the average of 5 runs. Moreover, 10\% of the train set is sliced for validation. Here are some details:
    \begin{itemize}
        \item \textbf{Data-types}: We have categorical and numerical features. For the CE-Fair Training part, numerical features of sensitive attributes are considered real-valued since they act as a regularizer and do not need to correspond to actual individuals. For the CE-Fair Prediction part, where we have counting over individuals sharing sensitive attributes, numerical features are considered to be discrete. For example, \emph{age} is considered a real-valued feature in training and an integer in prediction. Numerical features of nonsensitive attributes are always considered real-valued.
        \item $\bm{\rho}$: As discussed in \ref{sec:trainingresults}, for CE-Fair training, we have $\rho = 100\%, 1\%, 2\%$ for German, IPUMS Adult, and Law School, respectively. However, for the final evaluation on the test set (i.e., all the tables in the paper), models have been evaluated on the full test set for all datasets except for IPUMS Adult where we keep the same $\rho$ as train for runtime considerations.
    \end{itemize}
    
    \item \textbf{Model Architecture}\\
    The ReLU neural network model used across all experiments is a fixed architecture of 3 hidden layers of width 16. This is a fairly complex model for the tabular datasets used in such scenarios.
    
    \item \textbf{Pre-training}\\
    To pre-train the initial model, we run a grid-search over learning rate ($10^{-2}$, $10^{-3}$, $10^{-4}$) and batch size ($64$, $128$). We train each configuration for 500 epochs and select the model with the best loss on the validation set. This is the case with all datasets except for Law School which is more tricky to train on; for that, we use a learning rate of 0.01, a batch size of 256, train for 100 epochs, and take the last epoch model to get the best initial pre-trained model. 
    
    \item \textbf{CE-Fair Training}\\
    We re-train the pre-trained model through CE-Fair Training as discussed in Section \ref{sec:training} with the same learning rate and batch size used for pre-training. Each model is re-trained for 50 epochs with mentioned $\rho$. 
    Finally, we take the Nadir point of perfect \emph{accuracy} (1) and perfect \emph{CE rate} (0) and take the model with minimum $\ell_2$ distance to this Nadir point w.r.t. its train metrics.
    
    % \item \textbf{CE-Rate: CE-Fair Prediction}\\
    % % To compute CE-Rate defined in Section~\ref{sec:predictionresults}, we need to c
    % \aish{isnt this the smart counting strategy? why didnt we do this on all datasets?}
    % \kia{that's true. I did this because I was also logging \emph{CE Count} for which it is necessary to look for \emph{all} CEs but we ended up not reporting that.}
    % Given a test sample, we search for \emph{all} counterexamples. There is another version in which we search for only $\ceil*{\frac{|A(\boldsymbol{x})|}{2}}$ counterexamples and use tighter bound on the initial MILP encoding. This is more efficient in terms of runtime but might adversely affect numeric stability. We use this version for IPUMS Adult.
\end{itemize}

\section{Additional Experiments}

\subsection{Comparison to the original \textit{LCIFR} \cite{ruoss2020learning}}
Here we report the results that are imported from the \textit{LCIFR} \cite{ruoss2020learning} paper with their original implementation. Our overlapping datasets are German and Law School so here we only report those two.

Table \ref{tab:feta_lcifr_orig} demonstrates these results. The results of \textit{LCIFR} were obtained from their corresponding paper. We can see that at the same level of accuracy, we can guarantee fairness while \textit{LCIFR} can only empirically demonstrate fairness. For the Law School dataset, the empirical unfairness for \textit{LCIFR} is significantly large. Note in this experiment, we use more sensitive features for the German dataset which results in a more complex counterexample space. 

\begin{table}[h]
        \centering
        \caption{Comparison of FETA and LCIFR (original implementation)%\cite{ruoss2020learning}.
    }
        \resizebox{\linewidth}{!}{\footnotesize{
        \begin{tabular}{l|l| c c}
        \toprule
        \textbf{Dataset} & \textbf{Approach} &  \textbf{Accuracy (\%)} &\textbf{CE Rate (\%)} \\
        \toprule
        German & NN\textsubscript{b} & $76.70 \pm 2.78$ & $29.80 \pm 2.15$\\
        &LCIFR \cite{ruoss2020learning} & $ 75.53$ & $0.0$\\
        &\feta (CE-Fair Train. + Pred.) & $75.30 \pm 3.65$ & $0.0 \pm 0.0$\\
        \midrule
        Law School & NN\textsubscript{b} & $82.72 \pm 0.19$ & $41.18 \pm 2.05$\\
        & LCIFR \cite{ruoss2020learning} & $84.4$ & $48.9$\\
        & \feta (CE-Fair Train. + Pred.) & $84.14 \pm 0.97$ & $0.0 \pm 0.0$ \\
        \bottomrule
        \end{tabular}}}
    
    \label{tab:feta_lcifr_orig}
\end{table}

\subsection{Counterexample-guided Counting vs. Naive Enumeration}

We compare the performance of Counterexample-guided Counting for fair prediction (CE-Fair) to a naive enumeration method using the IPUMS Adult dataset, which has the most extensive range of sensitive features.

\begin{table}[h]
        \centering
        \caption{Comparison of CE-guided Counting and Naive Enumeration in runtime%\cite{ruoss2020learning}.
    }
        \resizebox{\linewidth}{!}{\footnotesize{
        \begin{tabular}{l|l| c c}
        \toprule
        \textbf{Dataset} & \textbf{Approach} &  \textbf{Average (s)} &\textbf{Total (s)} \\
        \toprule
        IPUMS Adult & Enumeration & $71.50$ & $655,655$\\
        &CE-guided Counting & $9.15$ & $83,905$\\
        \bottomrule
        \end{tabular}}}
    
    \label{tab:enum}
\end{table}

%In Table \ref{tab:all}, we summarize all the tables in the paper into one table where the effect of each step of the pipeline is clear.
%\input{table/results}

% \section{Assets}

% We will make the code publicly available; however, for the sake of the peer-review process, we have attached the code within an anonymous package along with the Appendix.

\section{Societal Impact} 

In this project, we focus on a specific fairness notion to verify, train and guarantee fair prediction of neural network models. However, we acknowledge that there is a huge literature on various notions of fairness, and fairness is context-dependent and should be defined relative to a task. Hence, our framework cannot and should not be used in every application domain. 

Although our approach can produce fair predictions, it is still based on a model produced by a machine learning algorithm. And it is important to note that FETA could suffer from the same disadvantages as the original model in aspects that we didn't consider in this work, such as privacy, explanation, safety, security, and robustness. Hence, the user must be aware of such a system’s limitations, especially when using these models to replace people in decision-making.

\section{Limitations} \label{sec:limits}
As discussed in \ref{sec:conclusion}, two limitations of FETA left for future work are i) adopting a fairness notion capable of capturing the relations among features, and ii) binding the counterexamples to follow the distribution of the data. Another issue, typical of approaches where such guarantees are provided, is scalability. Every year, state-of-the-art neural networks grow in size with a large number of parameters which poses incredible challenges for constraint-based verification approaches. Although the neural network model used in our experiments is fairly complex for tabular data, this approach might not scale to very deep networks. Indeed, it would be an interesting direction to address and explore the limits of scaling the model architecture. For example, future work could study how to modify the neural network learning algorithms to enable scalable constraint-based analysis. Finally, solvers tend to use floating-point approximations leading to numeric instabilities. Problem-specific solutions to make the approach more numerically stable could be a potential future work as well.

\end{document}